\documentclass{article}

\usepackage{arxiv}

\usepackage[utf8]{inputenc} 
\usepackage[T1]{fontenc}    
\usepackage{hyperref}       
\usepackage{url}            
\usepackage{booktabs}       
\usepackage{amsfonts}       
\usepackage{nicefrac}       
\usepackage{microtype}      
\usepackage{lipsum}
\usepackage{graphicx}
\graphicspath{ {./images/} }
\usepackage{xcolor}
\usepackage{hyperref}
\usepackage{breakcites}
\usepackage{amssymb}
\usepackage{subcaption}

\title{MVIP - A Dataset and Methods for Application Oriented \underline{M}ulti-\underline{V}iew and Multi-Modal \underline{I}ndustrial \underline{P}art Recognition}

\author{
 Paul Koch \\
  Fraunhofer IPK\\
  Pascalstraße 8-9\\
  10587 Berlin, Germany \\
  \texttt{paul.koch@ipk.fraunhofer.de} \\
   \And
 Marian Schlüter \\
  Fraunhofer IPK\\
  Pascalstraße 8-9\\
  10587 Berlin, Germany \\
  \texttt{marian.schlueter@ipk.fraunhofer.de} \\
  \And
 Jörg Krüger \\
  Technische Universität Berlin\\
  Pascalstraße 8-9\\
  10587 Berlin, Germany \\
  \texttt{joerg.krueger@tu-berlin.de} \\
}

\begin{document}
\maketitle
\begin{abstract}
We present MVIP, a novel dataset for multi-modal and multi-view application-oriented industrial part recognition. Here we are the first to combine a calibrated RGBD multi-view dataset with additional object context such as physical properties, natural language, and super-classes.
The current portfolio of available datasets offers a wide range of representations to design and benchmark related methods. 
In contrast to existing classification challenges, industrial recognition applications offer controlled multi-modal environments but at the same time have different problems than traditional 2D/3D classification challenges. 
Frequently, industrial applications must deal with a small amount or increased number of training data, visually similar parts, and varying object sizes, while requiring a robust near 100\% top~5 accuracy under cost and time constraints.
Current methods tackle such challenges individually, but direct adoption of these methods within industrial applications is complex and requires further research. 
Our main goal with MVIP is to study and push transferability of various state-of-the-art methods within related downstream tasks towards an efficient deployment of industrial classifiers. Additionally, we intend to push with MVIP research regarding several modality fusion topics, (automated) synthetic data generation, and complex data sampling -- combined in a single application-oriented benchmark.
\textbf{Github:} \textcolor{pink}{https://github.com/tbd}\\
\end{abstract}

\section{Introduction}

\label{sec:intro}
\begin{figure*}[h]
\begin{center}
\includegraphics[width=\textwidth]{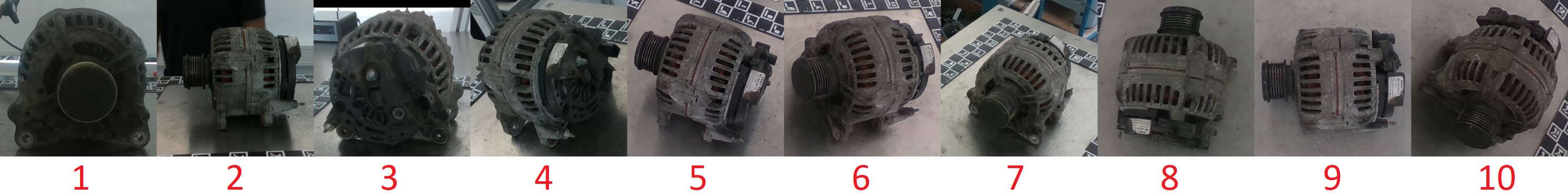}
\end{center}
\caption{An indexed ROI cropped MVIP image set featuring ten simultaneously captured Views}
\label{fig:sequence}
\end{figure*}

Vision-based classification systems have a broad range of industrial applications, e.g., 1) the intensification and sorting of incoming components into a warehouse; 2) the quality inspection and automated documentation of kitting and packaging processes; 3) key component identification of a broken machine to rapidly locate a fitting replacement within the warehouse. In reverse logistic, vision-based classification systems are used to identify old car components and classify them for remanufacturing, helping to reduce their relative carbon footprint~\cite{SCHLUTER2023414}. Ever since the success of AlexNet~\cite{AlexNet} in $2012$ related work for Vision-based classification has been developed with the ImageNet~\cite{ImageNet} classification benchmark\cite{VGG,ResNet,DenseNet,EffNet,ImageTransformers,Swin,Swinv2}. Due to these milestones, data-driven image processing heuristics have found their way into industrial applications, increasingly replacing traditional image processing approaches \cite{Shahrabadi_2022,SCHLUTER2021300,SCHLUTER2018384,app10103443}. Especially, it is noticeable that small problems with limited resources can thrive from transfer learning via publicly available pre-trained weight. A survey by Mazzei\&Ramjattan~\cite{Survey_onlyCNNs} ($2022$) report a general increase in Vison-AI related publications since $2016$. Here, they identify most work to be grounded on CNN based AI-architectures (e.g. ResNet) and encourage research towards the accessibility of state-of-the-art (SOTA) AI methods for easier adoption. However, a survey by Bertolini et al.~\cite{Rare_AI_Industry} ($2021$) finds that AI methods are limited to small groups of large international companies. Hence, small businesses thrive from accessible and adoptable AI-methods, while large companies can afford to push the SOTA towards their needs. Regarding vision-based classification, most public research focuses on single-view image classification. Although due to the nature of larger industrial objects (e.g., see Fig.~\ref{fig:sequence}) and the high similarity within objects, it is occasionally impossible to derive the correct object class from a single view. Therefore, we investigate within this work the SOTA for multi-view classification systems in order to identify their transferability towards our application oriented industrial part recognition benchmark (MVIP). 

\begin{figure}[h]
\begin{center}    
\includegraphics[width=8.5cm]{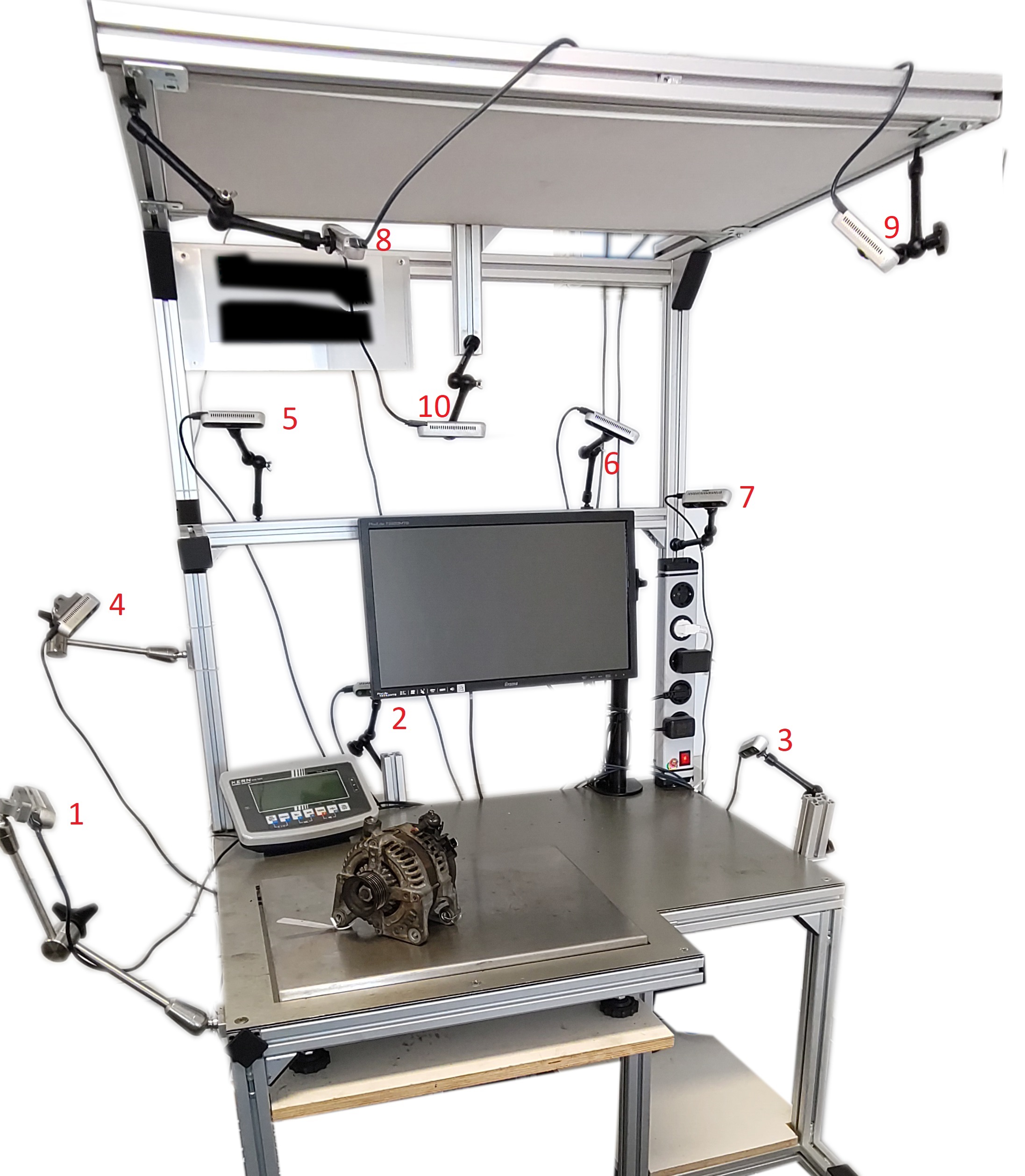}
\end{center}
\caption{Digitisation station used for MVIP}
\label{fig:table}
\end{figure}

With MVIP we contribute a novel multi-view (MV) and multi-modal (MM) dataset. For data acquisition, we designed a digitization and recognition station (see Fig.~\ref{fig:table}) as one could expect to be found on site in an industrial application for part recognition. The station is equipped with a vast set of task-relevant sensors (ten calibrated RGBD cameras and a scale) in order to investigate; A) which data benefits the industrial part recognition, and B) how to design and efficiently train a robust MV and MM model for industrial part recognition. In addition to the color, depth, and weight modalities, other modalities such as package-size (width, height, length), natural language tags (descriptions), and super-classes (general class spanning a common subset of classes, e.g. tool, car-component, etc.) are available in the dataset. This allows further research regarding modality-fusion and training or sampling methods. The calibrated MV-dataset allows 3D reconstruction of scenery and objects, which enables research regarding (automated) synthetic data generation and 3D based object recognition.

Our contributions with MVIP are manifold: 1) a novel multi-view (MV) and multi-modal (MM) dataset for application-oriented industrial part recognition; 2) a set of baseline investigations for MV and MM industrial object recognition; 3) a novel auxiliary loss for MV classification tasks; 4) Transformer-based MV-fusion; and 5) a novel approach for conditional MV-decoding for part recognition. With MVIP, we want to narrow the gap between related basic research and real industrial ML applications. 

\section{Related Works}

\textbf{Multi-View:} A recent ($2021$) survey on image fusion techniques~\cite{fusion_survey} identifies MV-fusion among other fusion techniques as an ongoing research topic, which is increasingly attracting more attention. Fusion can happen at different stages in a model architecture. Related works on MV-fusion employ a single image encoder to transform a set of images into vector-space (view tokens) and apply late fusion techniques \cite{mvcnn,View-GCN,MVTN,GVCNN,GIFT,MVFusionNet,RotationNet,MLVCNN,concat1,concat2,concat3,Multi-View_ML,sum_fuse}. The zoo of related SOTA late fusion techniques can be grouped into non-trainable, node-wise view-weighting ($\odot$), intra-view aware ($\leftrightarrow$), and inter-view aware ($\updownarrow$) methods. Consider tokenized view embeddings $\chi \in \mathbb{R}^{IxJ}$, where $I$ is the number of views and $J$ the number of hidden nodes, then $\odot = \forall_{j=1}^{J} \Sigma_{i=1}^{I} \chi_{ij}\nu_{ij}$, $\updownarrow = \Sigma_{j=1}^{J} \digamma(\chi_j)$, and $\leftrightarrow = \Sigma_{j=1}^{J} \forall_{I=1}^{I} \digamma(\chi_i)$. $\digamma$ is a trainable function and $\nu$ is a scalar found by some implementation of $\digamma(\subset \chi)$. E.g. pooling methods~\cite{mvcnn} are non-learn-able node-wise view-weighting methods for view aggregation. Likewise, convolution and fully-connected layers can be used to train a inter-view aware node-wise view-weighting policy. Otherwise, methods such as Squeeze-and-Excitation~\cite{Squeeze} (S.\&E.) use the individual embedding to determine a node-wise weighting, allowing an intra-view aware view aggregation. Methods based on concatenation of view embeddings~\cite{sum_fuse,MLVCNN,View-GCN} are inter-view and intra-view aware. Recent SOTA methods for View-Fusion have success with trainable~\cite{View-GCN} and non-trainable~\cite{MVTN} view aggregation methods.

\textbf{RGBD:} Unlike MV-fusion, SOTA methods for RGBD-related downstream tasks also employ hybrid fusion, where color and depth signals are gradually fused downstream. The work related to hybrid fusion can be grouped into methods that gradually fuse depth information with color information ($d \rightarrow c$)~\cite{RedNet} and a bi-directional fusion $c \leftrightarrow d$~\cite{CMX-rgbdfusion,Symmetric-Cross-modality-Residual-Fusion}. Depth directed fusion ($c\to d$) appears to be unnoticed in related work. Other work for RGBD-fusion-based downstream tasks employ late-fusion-based methods~\cite{Tokenfusion,RGBDfusion_old,RGBD-pretrained,DenseFusion,MVFusionNet}. 

\textbf{Transformers:} Ever since the introduction of Transformers~\cite{AttentionIsAll} into vision problems~\cite{ImageTransformers} they push the SOTA within vision-based downstream tasks~\cite{Swin,Swinv2,Detr,Deformable-Detr,Dino,segformer}. Due to their universal capabilities, transformers are found to be well suited to fuse information from different modalities~\cite{Clip,stable-diffusion,Tokenfusion,CMX-rgbdfusion,omnivore,omnivec,omnivec2}. Here, Omnivore~\cite{omnivore} and Omnivec~\cite{omnivec,omnivec2} use token-based modality fusion to reach state-of-the-art results on RGBD-based scene classification on SUN-RGBD~\cite{RGBD-SUN}. Therefore, we investigate the usage of Transformer-based methods for MV-fusion within industrial applications for part recognition. 

\textbf{MM \& MV DataSets:} Datasets for 6D-object-pose-estimation~\cite{PoseCNN,LineMod,t-less,GraspNet}, RGBD-segmentation~\cite{RGBD-SUN,RGBD-NYUv2,RGBD-ScanNet,kitti} and RGBD-instance-detection~\cite{PoseCNN,kitti,GraspNet,LineMod} drive RGBD based research within their related downstream tasks. Within MV and classification problems it is a common method within the field of 3D-part-recognition to render a set of 2D views from 3D parts and use image encoders to adopt MV-fusion for a combined classification~\cite{mvcnn,View-GCN,MVTN,GVCNN,GIFT,MVFusionNet,RotationNet,MLVCNN}. This research results in a vast set of synthetic~\cite{synt_ABC,synt_MCB,synt_modelnet,synt_partnet,synt_shapenet} and real world~\cite{real_CO3D,real_PASCAL3D+,real_video_amt,real_video_freiburgcars,real_video_GoogleScanned,real_video_MV-RGBD2011,real_video_objectron,real_video_scanobject} RGB(D) datasets for MV-fusion and classification investigations. Real-world MV datasets frequently use video-based capturing methods. Thus, MV can be sampled from the video. MV-RGBD~\cite{real_video_MV-RGBD2011} uses a turn table to capture objects from multiple fixed view points, while FewSOL~\cite{real_fewShot} and GraspNet~\cite{GraspNet} use a robot to capture object(s) from multiple view points. Albeit the use of synthetic industrial components~\cite{synt_ABC,synt_MCB} and fixed view points, none of the available datasets addresses a realistic industrial application for part recognition. Moreover, recent work within machine vision leverages the combination of images with other modalities such as natural language (NL)~\cite{Clip,stable-diffusion}. With MVIP, we are the first to our knowledge to bring the physical properties of objects, natural language, and MV-RGBD images into a single benchmark for application-oriented industrial part recognition.

\section{Methods}

\begin{figure}[h]
\begin{center}    
\includegraphics[width=8.5cm]{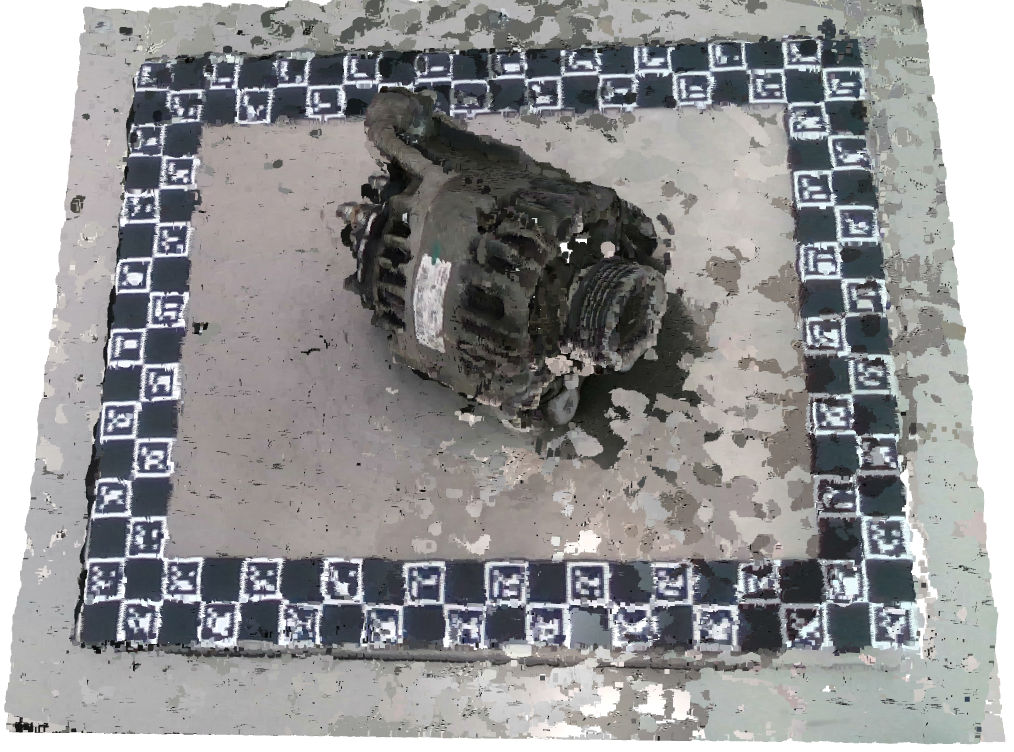}
\end{center}
\caption{Simple 3D-Object-Reconstruction given a single image set from MVIP.}
\label{fig:reconstruction}
\end{figure}

\textbf{The MVIP dataset} is captured on a digitization station as illustrated in Fig.~\ref{fig:table}. Ten RGBD cameras are mounted on the table construction, all facing a common point on the integrated scale. An ArUco-Board is surrounding the scale (see Fig.~\ref{fig:reconstruction}), thereby a 6D-Pose can be determined for each camera at any given time, given the camera's intrinsic parameters. Thus, the cameras are calibrated to each other within each captured image set (see Fig.~\ref{fig:sequence}). Due to the arrangement of camera perspectives and calibration, one set of images covers most of the object surface and allows 3D construction of the objects (Fig.~\ref{fig:reconstruction}). 
The front of the table is not equipped with cameras to provide space for a worker.\\
\indent Each object class featured in the dataset is rotated during digitization $12$ times (approximately $30^{\circ}$ steps), given a subjective "natural laying" position (see Fig.~\ref{fig:table} for illustration). Since some objects have multiple "natural laying" positions, $12$ rotations are repeated according to the subjective assessment of the worker. For test ($5$) and validation ($5$) purposes, ten additional image sets are captured, where the worker randomly moves the object on the scale according to a natural laying position. 

\begin{figure*}[h]
\begin{center}
\begin{subfigure}[t]{0.23\textwidth}
         \centering
         \includegraphics[width=\textwidth]{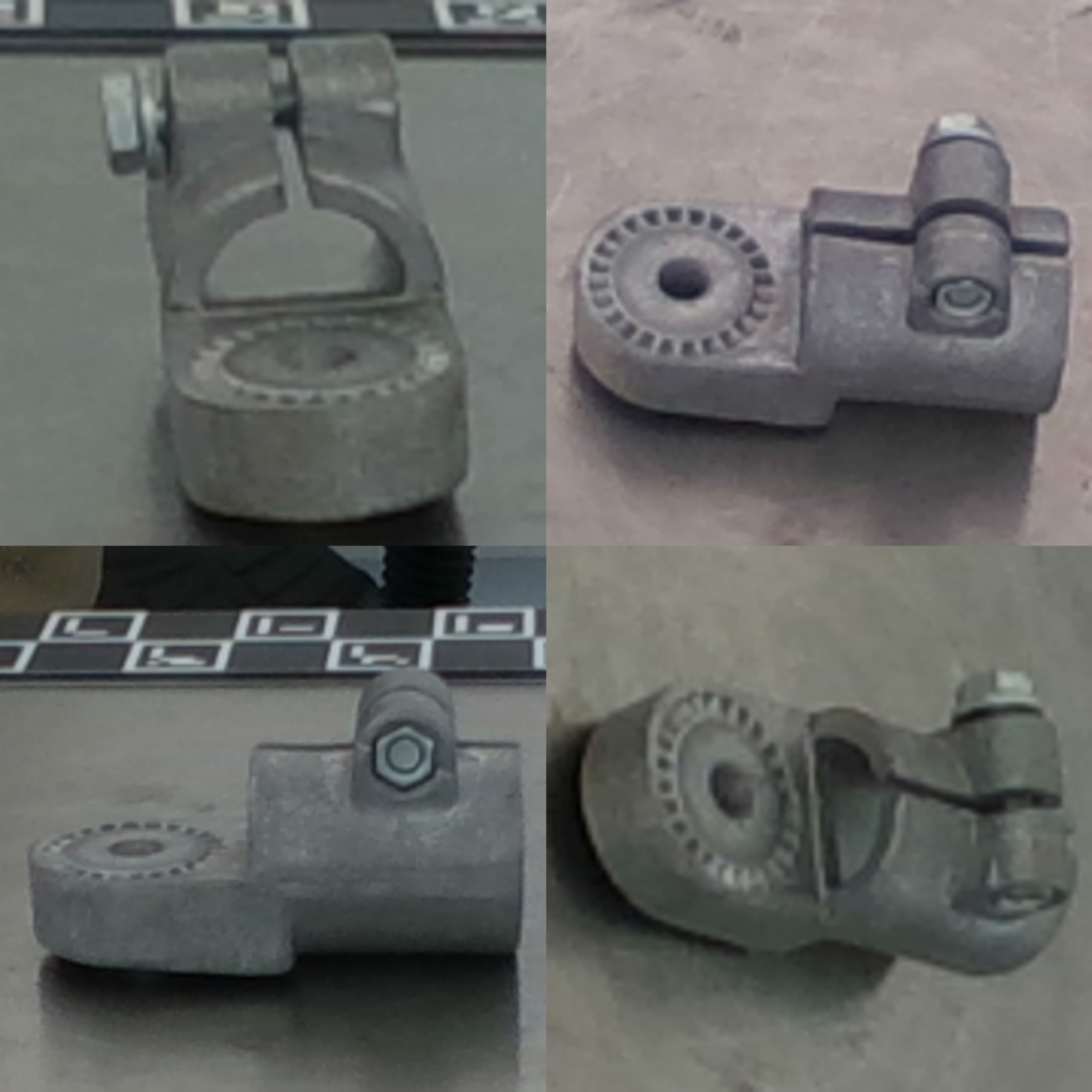}
     \end{subfigure}
    \begin{subfigure}[t]{0.23\textwidth}
         \centering
         \includegraphics[width=\textwidth]{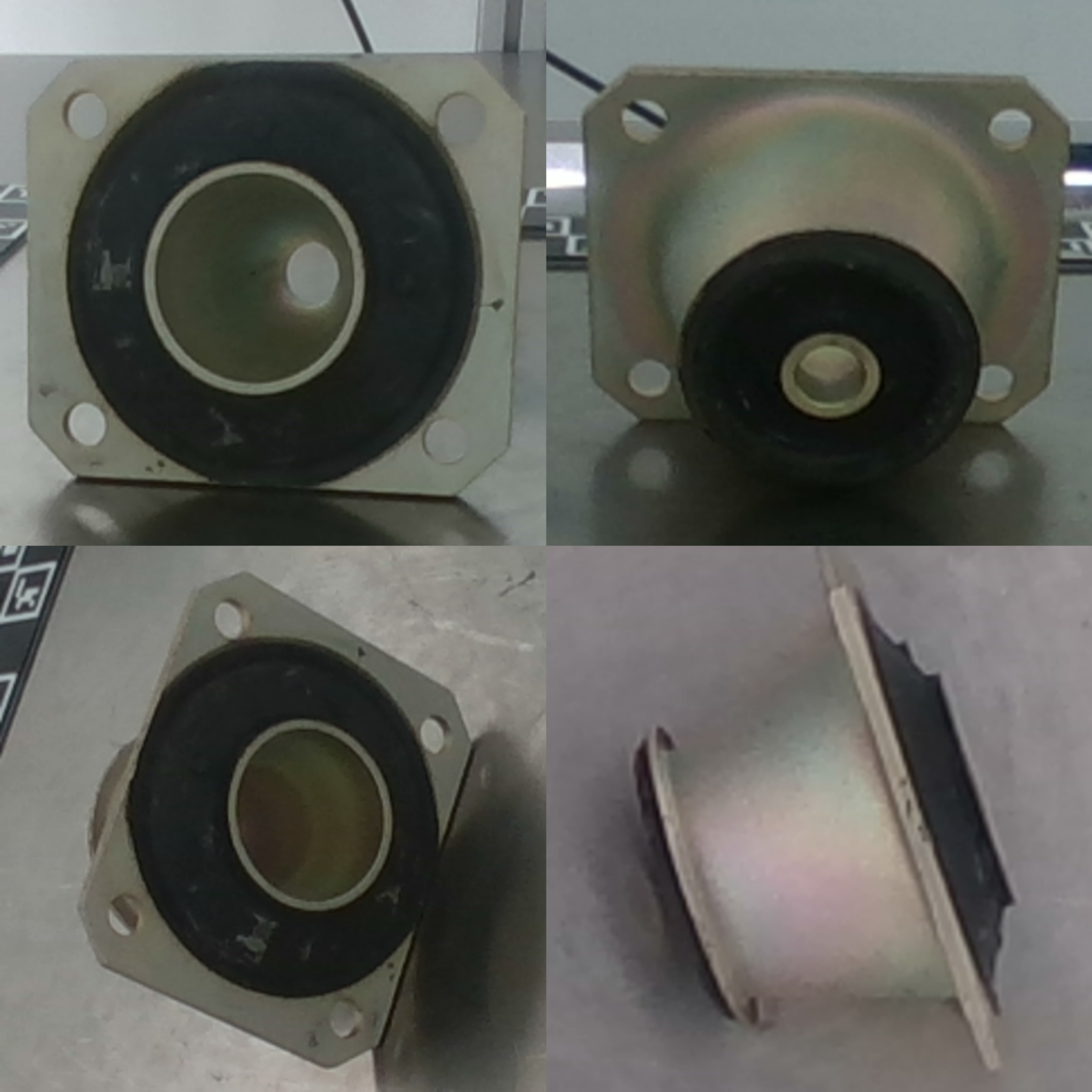}
     \end{subfigure}
    \begin{subfigure}[t]{0.23\textwidth}
         \centering
         \includegraphics[width=\textwidth]{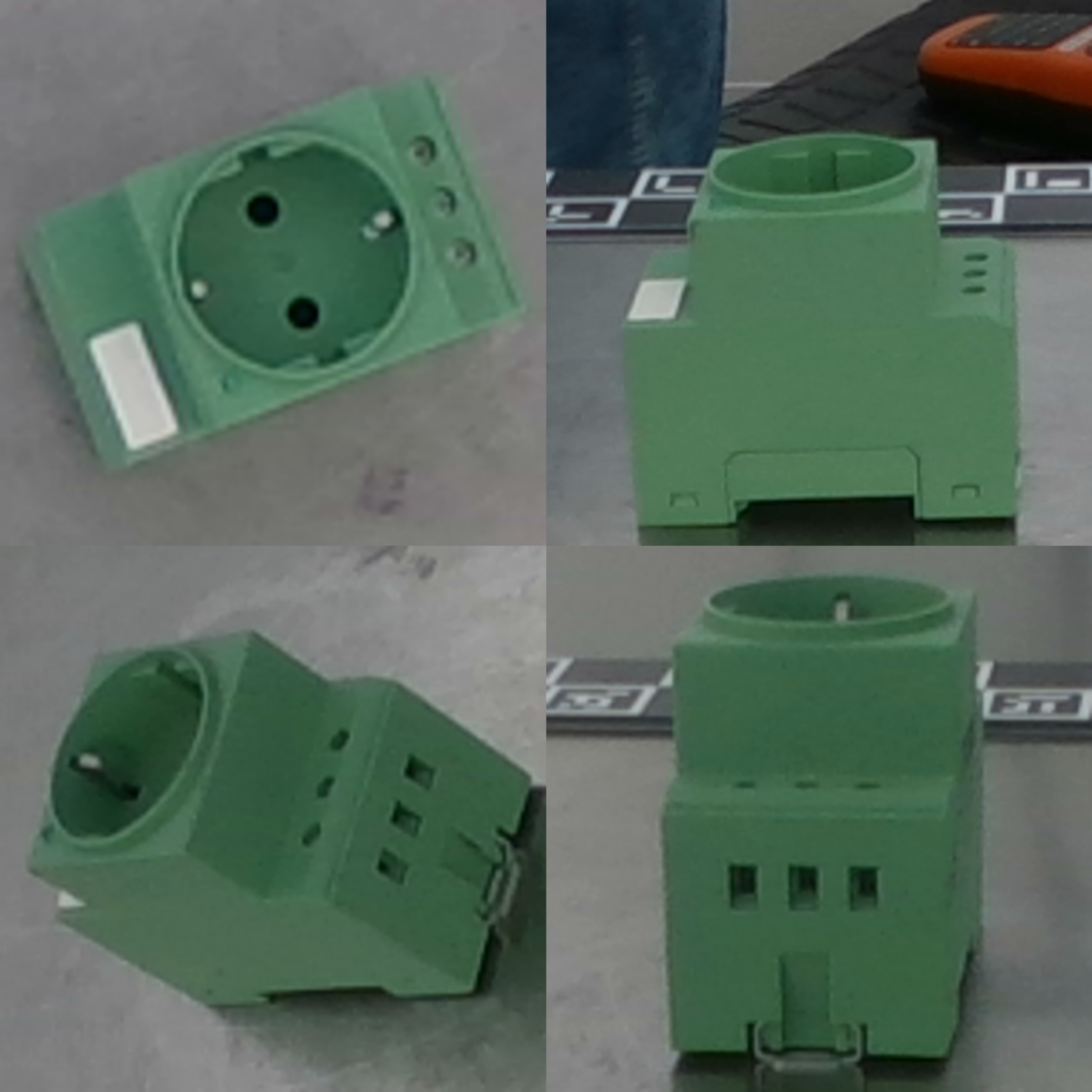}
     \end{subfigure}
     \begin{subfigure}[t]{0.23\textwidth}
         \centering
         \includegraphics[width=\textwidth]{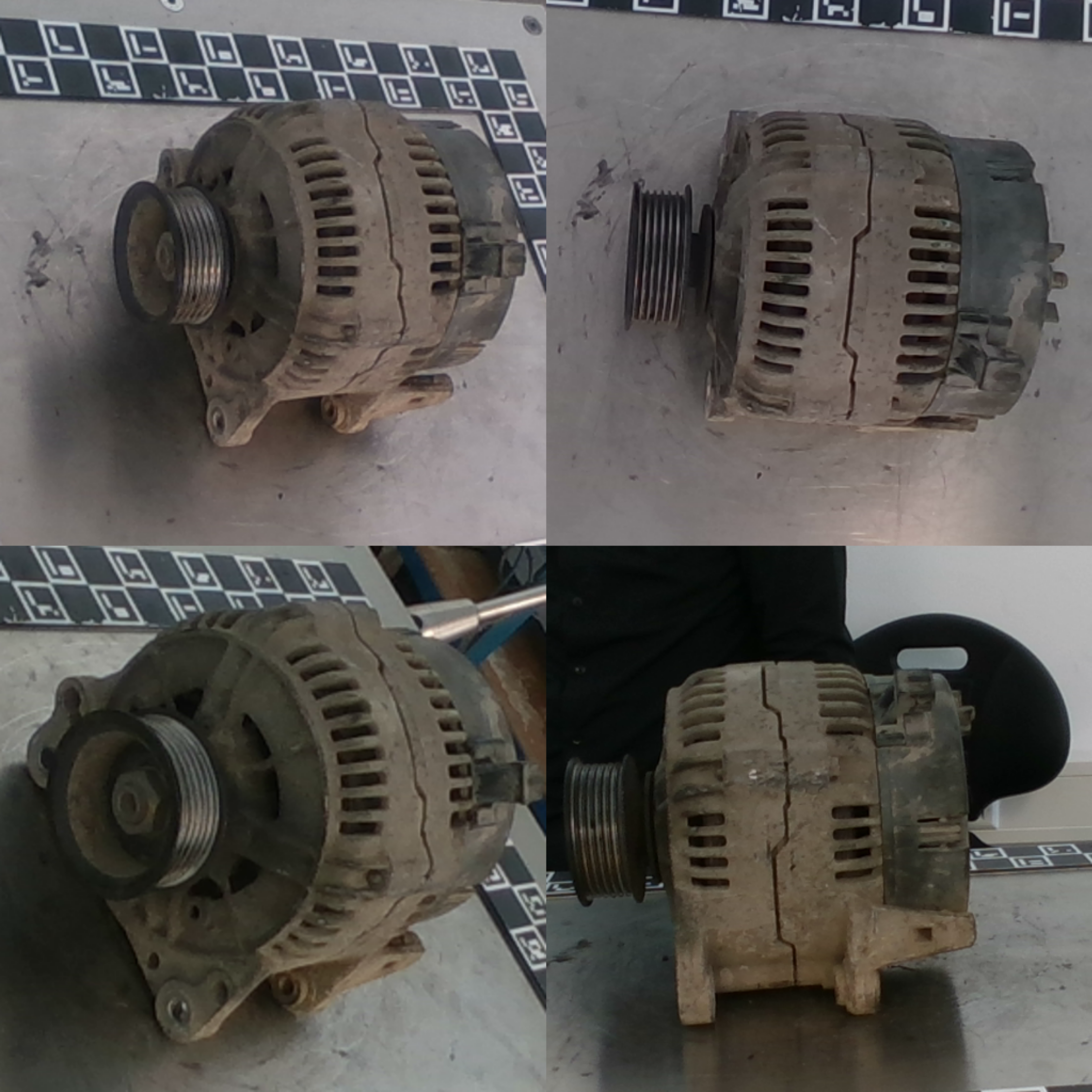}
     \end{subfigure}
\end{center}
\caption{Subset of industrial parts featured in the MVIP dataset. } 
\label{fig:multiple_objects}     
\end{figure*} 
Fig.~\ref{fig:multiple_objects} illustrates a subset of objects featured in the dataset. In addition to image data, the dataset features meta-data for all objects; weight, package size (length, width, height), object class, super-classes (general class spanning a common subset of classes), natural language (NL) tags, and generated view-wise object segmentation masks (thus, also ROI Bounding Boxes). For the segmentation masks, we annotated a small subset of 5\% and finetuned a segmentation model (\cite{dinov2}) to generate the remaining masks. This works well since the segmentation task is rather easy. The calibrated MV RGBD design of the dataset enables anyone to employ methods for 3D-Object-Point-Cloud and 3D-Scene-Reconstruction, 6D-Object-Pose Estimation, and Synthetic-Data generation. In total, MVIP features $308$ classes of industrial components (e.g. hammer, generator, camera adapter), which are grouped into $18$ super-classes (e.g. tools, car components, metal part). From eight categories (shapes, colors, materials, textures, conditions, size, weight, and density), MVIP uses $77$ NL-Tags to describe the classes. These tags describe in natural language the objects' conduction (diry, rusty, clean, etc.), shapes (round, sharp, edgy, pointy, etc.), and other visual attributes. MVIP has a total of $\approx570$ k images (all with resolution of $1280\times720$), while $71.276$ are RGB images. Each RGB image has the corresponding counterpart images: depth, HHA (following \cite{HHAorigin}), mean-RGB (mean images are averaged temporally over $1~sec$ for more stable data), mean depth, and segmentation (mask of the industrial part). Additionally, each image set is associated to a specific background set (scene without the industrial part). The $\approx282$ k images are available for training, while the $\approx108$ k images are used for validation and the other $\approx108$ k images for test. The industrial components featured in MVIP are set to be at least the approximate size of a fist (in a subjective assessment of the worker) and the maximum $350~mm\times450~mm\times300~mm$ with weight of $<15~kg$.

\begin{figure}[h!]
\begin{center}    
\includegraphics[width=\textwidth]{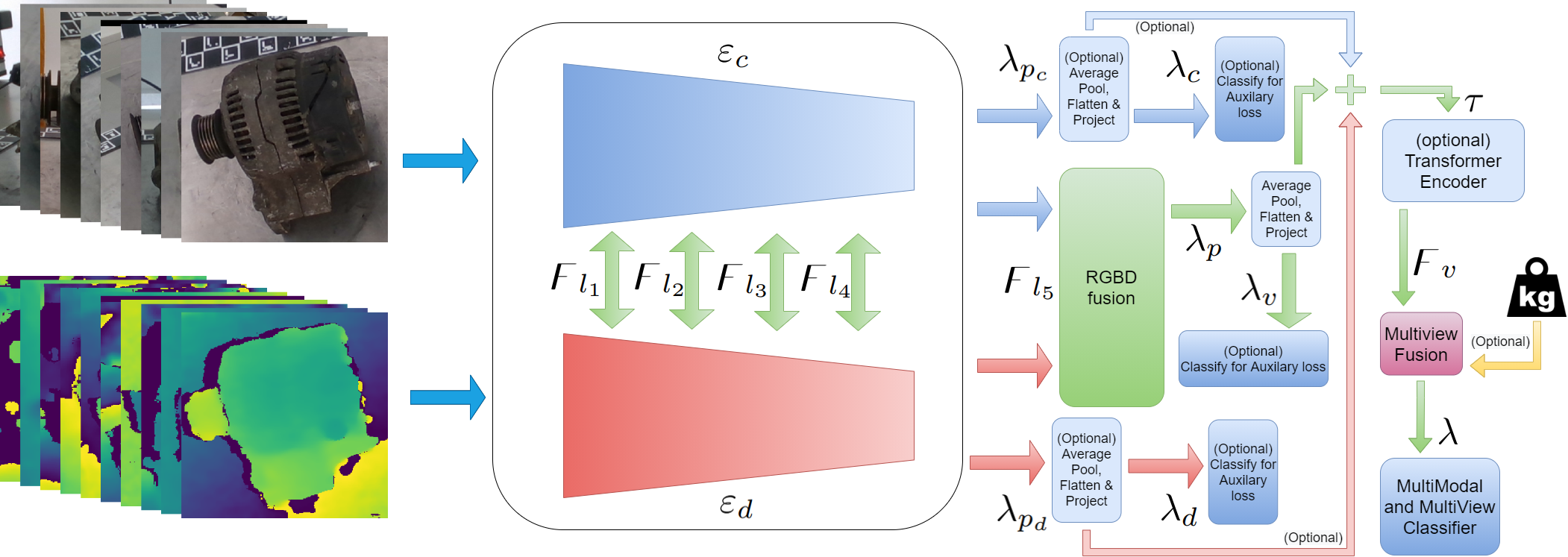}
\end{center}
\caption{Multi-view architecture used to train and evaluate our experiments, please find further description within our methods.}
\label{fig:MV-RGBD-Arch}
\end{figure}

\textbf{Architecture:} Inspired by related work, we created a modular training architecture (see Fig.\ref{fig:MV-RGBD-Arch}) in order to study ML-based industrial part recognition with MVIP. Our architecture uses a pre-trained CNN backbone~\cite{ResNet,EffNet}. Given the views $V$, the color and depth encoders $\epsilon_{c | d}$ project their inputs $\forall^V_{v=1}~c_v | d_v \in \mathbb{R} ^{Ch_{c | d}, H, W} \to \forall^V_{v=1} \forall^I_{i=1}~l_{c | d, v, i} \in \mathbb{R}^{Ch_{emb_i} \times H / {2^i} \times W / {2^i}}$. \\
\indent In our implementation, we use the ResNet50~\cite{ResNet} encoding architecture $\epsilon_{c | d}$ where $I=5$. At every embedding stage $l_{c|d, v, i} = \epsilon_{c|d, i}(\hat{l}_{c|d, v, i-1})$ a RGBD-fusion $\digamma_{l_i}$ combines the embeddings $l_{c \land d, v, i} \to \hat{l}_{c | d, v, i}$, where $\hat{l}_{c | d, v, 0} = c_v | d_v$ and $\digamma_{l_i}$ is either a one-directional fusion $x = c | d;~\hat{l}_{x, v, i},~\hat{l}_{\neg x, v, i}  = \digamma_{l_i}(l_{c \land d, v, i}),~l_{\neg x, v, i}$ or bi-directional fusion $\hat{l}_{c, v, i},~\hat{l}_{d, v, i} = \digamma_{l_i}(l_{c \land d, v, i})$. \\
\indent The final fused embedding $\hat{l}_{v, I} = \digamma_I(l_{c \land d, v, I}) \in \mathbb{R}^{Ch_{emb_I} \times H/{2^I} \times W/{2^I}}$ is passed through an average pooling layer followed by a flattening before being projected via the fully connected layer $\lambda_p$ into the view embeddings $\chi_v \in \mathbb{R}^{1 \times Ch_V}$.
Optionally, the non-fused embeddings $l_{c | d, v, I}$ are also passed through an average pooling layer followed by a flattening and projected via a fully connected layer $\lambda_{c | d}$ into the auxiliary output $O_{v, c | d} \in \mathbb{R}^{1 \times N}$, where $N$ is the number of classes. Likewise, $\chi_v$ is projected by $\lambda_v$ into $O_v \in \mathbb{R}^{1 \times N}$. The view embeddings $\chi_v$ are stacked for a combination of embeddings $\chi = cat(\forall^V_{v=1} \chi_v)$. An optional transformer encoder $\tau$~\cite{AttentionIsAll} can be applied in $\chi$ to employ self-attention between all $\chi_v$. 

Eventually, we use MV-fusion $\digamma_v$ to reduce the dimension of $\chi \in \mathbb{R}^{V \times {Ch_v}} \to~\in \mathbb{R}^{1 \times {Ch_v}}$ before a final fully connected layer $\lambda$ projects $\chi \to 0 \in \mathbb{R}^{1 \times N}$. Optionally, our transformer-based implementations for $\digamma_v$ additionally take the object weight in kg as input, which is used as an anchor (conditional decoding~\cite{stable-diffusion}). Likewise to Positional Encoding~\cite{AttentionIsAll} we encode the object weight into a sequence of cosine and sine frequencies and add the vector to the decoding signal of our transformer-based MV-fusion implementations. For RGB-MV classification, the modules $\epsilon_d$ and $\forall^{I-1}_{i=1} \digamma_{l_i}$ are removed, while $\digamma_{L_I}$ is an identity function and $\forall^I_{i=1}~\hat{l}_{c, v, i} = l_{c, v, i}$.

\textbf{MV-fusion implementations: } In our experiments, we study different MV-fusion types within MVIP and effect of scaling the trainable parameters. We summarize the implementations in our investigations in Tab.~\ref{tab:fusion}.
These methods (in addition to $\tau$) are implementations of $\digamma_v$ and are used to reduce the tokenized embeddings of the views $V$. Following related work, we use the pooling aggregation introduced by~\cite{mvcnn} in 2015 and is still used recently~\cite{MVTN}. Following Feng et al.~\cite{GVCNN} we implement a Conv.-based node-wise view aggregation which gradually fuses the views in sequential layers. However, most related work on view aggregation is designed for problems with $12$ and $20$ views for 3D object recognition~\cite{View-GCN,MLVCNN,RotationNet}. E.g., Wei et al.~\cite{GVCNN} uses a graph-based view aggregation implementation which is only suitable for views $V$ where $V/4 > 1~\&~V/4 \in \mathbb{N}$. Regarding Transformers ($\tau$, Tr-En, Tr-EnDE) we are the first to our knowledge to bring the attention mechanism to tokenized view aggregation. Here $\tau$ is a simple Transformer encoder, introduced by Vaswani et al.~\cite{AttentionIsAll}. The Transformer encoder-decoder (Tr-EnDe)~\cite{AttentionIsAll} is used by Carion et al.~\cite{Detr} to extract information from encoded image information through a trainable decoding query and cross-attention. We adopt this approach for view aggregation by decoding class information with a single trainable query from the tokenized view embeddings. Dosovitskiy et al.~\cite{ImageTransformers} append a trainable decoding query to the tokenized embeddings before forwarding it through the Transformer encoder (Tr-En)~\cite{AttentionIsAll}. Afterwards, a final classification is done only on the basis of the output at the index of the trainable query. Thus, the attention mechanism is applied while reducing the number of trainable parameters compared to the encoder-decoder approach. The transformer-based view aggregation methods are intra- and inter-view aware, since the attention mechanism can attend to every node. Furthermore, we adopt the Squeeze-and-Excitation Networks (S\&E) introduced by Hu et al.~\cite{Squeeze} as a comparison for intra-view-aware view aggregation. Here, each tokenized view embedding is attending to every node within the view and used to determine a node-wise scalar, which is used for a simple sum view aggregation. In our shared-S\&E (S.S\&E) we use a single module to compute the scalars for every view, rather then having a individual S\&E module for every view. An overview of the implemented fusion methods can be seen in Tab.~\ref{tab:MV-fusion-implementations}.
\begin{table}[h!]
  \begin{center}
  \begin{tabular}{|lccc|}
    \hline
    MV-fusion & Types & Params.& Blocks / Layers\\    
    \hline \hline
     Max Pool~\cite{mvcnn}& $\odot$ & - & -\\
     Mean~\cite{mvcnn} & $\odot$ & - & -\\
     Conv.~\cite{GVCNN} & $\odot \updownarrow$ & 13 & 3 Layers\\
     S{\&}E~\cite{Squeeze} & $\odot \leftrightarrow$& 0.4Mio& 1 Block\\
     S.S{\&}E~\cite{Squeeze} & $\odot \leftrightarrow$ & 0.13Mio& 1 Block\\
     MLP &$\updownarrow \leftrightarrow$ & 13Mio& 3 Layers\\
     Tr-En~\cite{ImageTransformers}& $\updownarrow \leftrightarrow$ & 8.4mio& 1 Block\\
     Tr-EnDe~\cite{Detr}& $\updownarrow \leftrightarrow$  & 21mio& 1 Block\\
     \hline
     $\tau$~\cite{AttentionIsAll} & $\updownarrow \leftrightarrow$ & 8.4mio & 1 Block\\
    \hline     
  \end{tabular}
  \end{center}
  \caption{Implementation details concerning the MV-fusion methods used in this work. We denote the MV-fusion types with $\odot$ for node-wise view-weighting, $\updownarrow$ for intra-view aware, and $\leftrightarrow$ for inter-view aware methods.}
  \label{tab:MV-fusion-implementations}
\end{table}

\textbf{Multi head auxiliary loss:} Inspired by related work~\cite{Inception,Detr,FeaturePyramidNet} we introduce a novel auxiliary loss for MV classification which is defined as;\\
\begin{equation}
    \zeta_v = \zeta_{cls}(o_v, y_v) 
\end{equation}
    
\begin{equation}
    \zeta_{v_{CD}} = \frac{1}{3}(\zeta_{cls}(o_v, y_v) + \zeta_{cls}(o_{v_c}, y_v) + \zeta_{cls}(o_{v_d}, y_v))
\end{equation}

\begin{equation}
    \zeta_{MH} = \frac{1}{V+1}( \zeta_{cls}(o, y_v) + \Sigma_{v=1}^{V} \zeta_v)
\end{equation}
\begin{equation}
    \zeta_{MH_{RGBD}} = \frac{1}{V+1}( \zeta_{cls}(o, y) + \Sigma_{v=1}^{V} \zeta_{v_{CD}})
\end{equation}
where $\zeta_{cls}$ denotes cross-entropy-loss, $y$ is the MV target class, $y_v$ is the view wise target class, $V$ is the number of Views, and $o$, $o_v$, $o_{v_c}$, and $o_{v_d}$ are the overall, full-view, color-view, and depth-view predictions, respectively. With our multi head auxilary loss (MH-loss) we embrace gradients view-wise within each encoding modality, thus forcing contributions from every view and modality, which hinders the classifier to specify/overfit on a certain view and modality. This is especially important if the pretrained encoder initially favors a certain modality or view.

\textbf{Weight-based Classification:} The object weight $\omega~[kg] \in \mathbb{R}$ is a one-dimensional non-unique property. Thus, classifying objects purely based on weight is ambiguous, since the probability that the objects share a common $\omega \pm e$ is large, where $e$ is the scale resolution. The scale use in MVIP has a constant weight error of only $0.002~kg$, but is heavily affected by offsetting objects from the scale center. Thus, any weight-based classifier must be robust to disturbances in the weight measurement. Following positional encoding~\cite{AttentionIsAll} (PE) we use a set of $d \in \mathbb{N}$ sine-cosine functions with varying frequencies to encode the weight $\omega$ into a vector of size $1\times 2d$. We Train a 4-Layer MLP to upscale the weight $1 \times 2d \rightarrow 1\times d_h$, where $d_h$ is the number of hidden nodes. Eventually, a fully-connected layer (FC) is used for classification. During training, we apply randomly a constant or proportional weight error uniformly sampled from the distribution $\pm 0.01~kg$ or $0.01\omega~[kg]$, respectively. We train PropertyNet using cross-entropy loss and Adam Optimizer for $10$k epochs with a scheduled cosine one-cycle learning rate between $10^{-7} \nearrow 10^{-6} \searrow 10^{-9}$ (max at 50\% training) and a batch size of $512$. 

\section{Experiments \& Discussion}
\textbf{Experiment design: }All experiments are conducted on the same machine ($2\times$Nvidia RTX 3090) with the same set of hyper-parameters for training. For fair comparison, the batch-size stays constant at $32$, which means that the largest model utilizes the 48GB GPU-RAM at most, while smaller models theoretically could have used a larger batch size. For a baseline establishment, the ResNet-$50$~\cite{ResNet} architecture with pre-trained ImageNet\cite{ImageNet} weights is used as an image encoder. We train our classifiers with our auxiliary loss and Adam optimization at an cosine one cycle~\cite{one-cycle} scheduled learning rate of $10^{-5} \nearrow 10^{-4} \searrow 10^{-6}$ (max at 50\% training). If not stated otherwise, we report the maximum observed accuracy out of five runs on MVIP for $50$ epochs with ROI crops, colorjitter, flip, rotation and random view order augmentations on three view RGB with a resolution of $224\times224$ pixels (ROI crops are up-sampled if needed). During testing, we fix the view order and also indexes if additional views are available during training. 

\indent \textbf{Stability and Resolution: } 
In Tab.~\ref{tab:stability} we investigate the stability of our proposed architecture w.r.t. 3-view-RGB and our MH-loss. Here we find our MH-loss to yield superior and more stable results compared to pure end-to-end training with cross-entropy-loss. 
\begin{table}
  \begin{center}
  \begin{tabular}{|lcccc|}
    \hline
    MV-fusion &  $\#~\tau$ & Params. & $\%$ & $\%_{MH}$\\    
    \hline \hline
     Max Pool& 0 &0 & 73.7$\pm$0.4 & \textbf{94.6$\pm$0.2}\\
     Conv. & 0 & 13 & \textbf{89.6$\pm$2.0}& 94.3$\pm$0.9\\
     Conv. & 1 & 8.4Mio & 66.7$\pm$4.9 & 91.2$\pm$0.8\\
    \hline
  \end{tabular}
  \end{center}
  \caption{Stability of our implementations w.r.t. three view RGB and trainable MV-fusion parameters (Params.). }
  \label{tab:stability}
\end{table}
In Tab.~\ref{tab:resolution} we inspect the effect of using a higher resolution, the usage of multi-scale inputs, and up-sampling of ROI crops. Despite our observations regarding the input resolution, we continue our experiments with an input size of $224\times224$ without multi-scale to keep a constant batch size in every experiment. 
\begin{table}
\begin{center}
  \begin{tabular}{| c c c c c|} 
    \hline
     Resolution & Multi-Scale & Up-Sampling & BS&$\%$ \\ 
    \hline \hline
     $224\times224$ & - & - &32&81.4\\ 
     $224\times224$ & - & \checkmark &32&82.7\\ 
     $224\times224$ &\checkmark & \checkmark &32 &88.8 \\ 
     $512\times512$ &- & - & 32&88.2\\ 
     $512\times512$ & - & \checkmark &32&\textbf{92.9}\\ 
     $512\times512$ & \checkmark & \checkmark &14&91.6\\ 
    \hline
  \end{tabular}
\end{center}
  \caption{ROI cropped 3 view RGB results concerning resolution. Multi-Scale varies the resolution $\pm 10\%$ while Up-Sampling scales the roi crop to the desired resolution instead of forcing a bigger ROI crop. $512\times 512$ with multi-scale has a decreased batch size in order to fit in our GPU-RAM.}
  \label{tab:resolution}
\end{table}

\indent \textbf{MV-fusion:} In Tab.~\ref{tab:fusion} we report the results concerning view aggregation. We find that our MH-loss yields superior results across the board. Interestingly we observe max pooling to lose performance with view increase (especially without our MH-Loss). This phenomenon might be explained by an accumulated self-reinforcing view specific feature extraction, which is more likely to lead to overfitting and only present for the Max pooling implementation. Our MH-loss circumvents this overfitting phenomenon due to additional view-wise backpropagation. As we employ random view ordering during training to reduce overfitting (see Tab.~\ref{tab:augs}), we observe the trainable node-wise weighing convolution to collapse towards average pooling. Moreover, we find that our classifiers tend to overfitting with increasing MV-fusion complexity and view numbers. Here we find that the lightweight intra-view weighing approach of S.S\&E yields marginal gains compared to non-trainable view aggregation. Transformer-based implementations are notoriously hard to train with a tendency to overfit~\cite{AttentionIsAll,ImageTransformers,Detr,Bootstrapping_ViTs}. Related work found self-supervised (SSL) pre-training to be key to the success of transformer-based implementations~\cite{Dino,Bert,transformers_hard_to_train}, which opens for further SSL-based investigation for tokenized view aggregation with transformers. Likewise, we find from the training logs that the trainable but not pre-trained view aggregation to be less efficient w.r.t. training time. Thus, pre-training is crucial for efficient training of complex view aggregation methods. We conclude that trainable inter view aware methods for view aggregation are not adding compared to non-trainable alternatives as the view order is invariant. However, we see the potential for complex intra-view \& inter-view aware methods, such as transformers, to yield better results if (pre)trained in larger datasets.
\begin{table*}
  \begin{center}
  \begin{tabular}{|lcccccccc|}
    \hline
    MV-fusion & Fusion-Types& Trainable& \underline{L}ayers or &$\tau$ & \multicolumn{2}{c}{$\%$} &\multicolumn{2}{c|}{ $\%_{MH}$}\\
    &&Parameters&\underline{B}locks&&2 Views&3 Views&2 View&3 Views \\    
    \hline \hline
     Max Pool\cite{mvcnn} &$\odot$&-&-&-& 88.2 & 74.4 & 95.3 & 94.7\\
     Mean\cite{mvcnn} &$\odot$ &-&-& - &\textbf{92.9}&87.9 &95.3&\textbf{95.5}\\
     Conv.\cite{GVCNN} &$\odot$&13&3l&-&91.6&\textbf{92.1}& 94.6&\textbf{95.5}\\
     Conv.\cite{GVCNN} &$\odot$&8.4mio&1B+3L&1&84.5&72.5 & 93.4&92.4\\
     S{\&}E\cite{Squeeze} &$\odot \leftrightarrow$&Views$\times$0.13mio&Views$\times$1L&-& 91.9 &84.2 &94.5&94.6\\
     S.S{\&}E\cite{Squeeze} &$\odot \leftrightarrow$&0.13Mio&1L& - & 91.2 &83.7&\textbf{95.6}&94.5\\
     MLP&$\updownarrow \leftrightarrow$&Views$\times$4.3Mio&3L& - & 86.2 & 80.5& 93.7&93.6\\
     Tr-En~\cite{ImageTransformers} & $\updownarrow \leftrightarrow$&8.4mio&1B&- & 74.44 &56.5 & 94.6&  92.0\\
     Tr-De~\cite{Detr} &$\updownarrow \leftrightarrow$&21Mio&1B & - & 88.4 & 78.1 &95.3& 94.6\\
    \hline
  \end{tabular}
  \end{center}
  \caption{Results concerning 2\&3 view fusion on MVIP. The number of $\tau$ indicates how many Transformer-Encoding blocks are used before MV-fusion (see Fig.~\ref{fig:MV-RGBD-Arch}). All experiments trained for $100$ epochs. The $\%_{MH}$ denotes results concerning classifiers trained with our MH-loss. We denote the MV-fusion types with $\odot$ for node-wise view-weighting, $\updownarrow$ for intra-view aware, and $\leftrightarrow$ for inter-view aware methods.}
  \label{tab:fusion}
\end{table*}

\begin{figure}[h!]
\centering
\includegraphics[width=8.5cm]{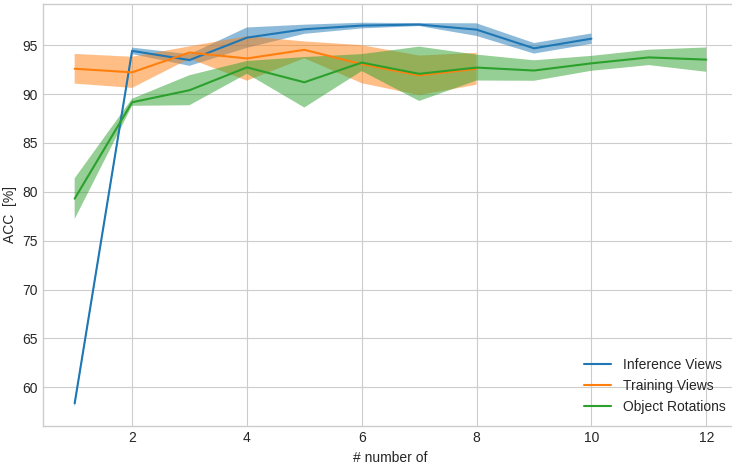}
\caption{\textbf{Views:} Results concerning increasing the number of input views included in the MV-architecture (Inference Views), number of views available during training of a 3 view RGB classifier, and the number of available training image sets (Object Rotations) for 3 view RGB classifier.}
\label{fig:view_rot_exp}
\end{figure}

\textbf{Saturation of Accuracy:} In Fig.~\ref{fig:view_rot_exp} we summarize our results regarding increasing the number of input views in our MV-architecture, available views within the training data, and the effect of reducing the number of training image sets. Here we observe the most significant performance gain already with two views, while reaching a saturation of accuracy at around five views. Further increasing the number of views does not significantly improve the performance w.r.t. MV nor the number of training samples. Similar to our MV-fusion related results (see Tab.~\ref{tab:fusion}), we observe a significant performance drop with three Views. It can further be observed that some training samples w.r.t view ID and rotation ID appear to corrupt the training and cause overfitting. We identify the view IDs two and seven as being especially affected by artifacts introduced by the worker. Moreover, we argue that certain view/rotation configurations emerge more overlapping and occurrences of artifacts which are affecting the statistical generalization.

\textbf{Augmentations} such as jittered ROI-crops, flips, and rotations are found to be default settings for training single-view image processing models in order to diversify the dataset and regularize the learning. However, it can be argued that using such methods for MV-approaches might hinder the model from "considering" the 3D nature of a given presented part. Investigations on the effect of keeping the 3D structure complete of any part during MV-training can be found in Tab.~\ref{tab:augs}. Here we observe that our methods within MVIP have a general problem with overfitting, even after background elimination via ROI crops. We find random ordering of views to be most important in order to hinder the classifiers to overfit.  
We also investigate shuffling views regarding class and view index in hope of better results due to a more generalized image encoder $\epsilon$ and less overfitting on fixed but order-invariant view sets. For inter-class shuffling we switch to binary-cross-entropy-loss in a pre-training stage 1 (30\%), and disable it in a stage 2 to train a classifier with cross-entropy-loss. Here we observe no further generalization; in fact, the performance decreases, which indicates that maintaining the image set structure is important for MV-classification. 
\begin{table}[h!]
\begin{center}    
  \begin{tabular}{|c ccccl|}
    \hline
    Crop & Flip & Rotate & Random View & Shuffle Views & $\%$\\    
    \hline \hline
     \checkmark & - & - & - & - & 64.0 \\ 
     \checkmark & \checkmark & - & - & - & 67.1\\ 
     \checkmark & - & \checkmark & - & - & 80.8\\ 
      - & \checkmark & \checkmark & - & - & 79.8 \\ 
     \checkmark & \checkmark & \checkmark & - & - & 82.7 \\ 
     \checkmark & \checkmark & \checkmark & \checkmark & - & \textbf{94.9} \\ 
     \checkmark & \checkmark & \checkmark & \checkmark & VW \& CW & 90.26 \\ 
     \checkmark & \checkmark & \checkmark & \checkmark & $\neg$VW \& CW & 92.0 \\ 
     \checkmark & \checkmark & \checkmark & \checkmark & $VW \& \neg$ CW & 84.3 \\ 
     \checkmark & \checkmark & \checkmark & \checkmark & $\neg$ VW \& $\neg$ CW & 81.7 \\ 
    \hline
  \end{tabular}
\end{center}
  \caption{Results concerning MV-augmentations and MV shuffling methods. View wise (VW) and class wise (CW) shuffling between MV image sets.}
  \label{tab:augs}
\end{table}
 One could try to use single-view pretraining rather than view shuffling for a better pretrained image encoder $\epsilon$, which we investigate further in Tab.~\ref{tab:pre-trained_exp}. In our implementation, we find a pre-training stage to be unfavorable. This could be due to a catastrophic forgetting of generalized weights from Imagenet pre-training before reaching the end-to-end MV classification training. 
\begin{table}[h!]
  \begin{center}      
  \begin{tabular}{| l |ccc l|} 
    \hline
    Tag & Train Views& Test Views & Pretrained on&$\%$ \\    
    \hline \hline
     A-SV & 1 & 1 &ImageNet&57.9\\
     B-SV & 3 & 1 &ImageNet&85.5\\
     C-SV & 10 & 1 &ImageNet&\textbf{88.4}\\
     \hline
     MV & 3 & 3 &ImageNet&\textbf{94.0}\\
     MV & 10 & 3 &ImageNet&93.6\\
     A-MV & 3 & 3 &A-SV & 62.2\\
     B-MV & 3 & 3 &B-SV & 91.4\\
     C-MV & 10 & 3&C-SV & 84.4\\
    \hline
  \end{tabular}
  \end{center}
  \caption{Results concerning pre-trained-weights for the image encoder $\epsilon$ w.r.t. available views to sample from during training and the number of test views.}
  \label{tab:pre-trained_exp}
\end{table}

\begin{table}[h!]
  \begin{center}
  \begin{tabular}{|lcccc|}
    \hline
    Input & Norm. & Fusion& $\%$ & $\%_{VCD}$\\    
    \hline \hline
    3RGB (Base) &-&-&\textbf{94.6}&-\\
    \hline
    HHA~\cite{HHAorigin}& -&-&44.2&-\\
    D& - &-&51.6&-\\
    D& \checkmark&-&47.2&-\\
    D$^\star$~\cite{RedNet}& - &-&46.2&-\\
    \hline
    3RGBD& - &$c\leftarrow d$ &84.0&-\\
    3RGBD& - &$c\leftrightarrow d$&81.1&-\\   
    3RGBD& - &$c \rightarrow d$& 93.6&93.8\\
    3RGBD& \checkmark &$c \rightarrow d$ & - &\textbf{94.5}\\
    3RGBHHA& - &$c \rightarrow HHA$ & 93.1&94.0\\
    3RGBD& \checkmark & - & 94.0&-\\
    3RGBD$^\star$~\cite{RedNet}&\checkmark & - & 93.6&-\\  
    \hline
  \end{tabular}
  \end{center}
  \caption{Results concerning 3 view RGBD-fusion trained for $200$ epochs with our RGBD MH-loss function and where $\%_{VCD}$ denotes the accuracy achieved when applying MV-fusion $\digamma_v$ on the concatenated embeddings from the RGBD \underline{v}iew, \underline{c}olor, and \underline{d}epth signals (see Fig.~\ref{fig:MV-RGBD-Arch}). Here, we use S.\&E.~\cite{Squeeze} for efficient baseline RGBD-fusion. We mark a $^\star$ where we use pre-trained depth encoding weights from~\cite{RedNet}. }
  \label{tab:RGBD}
\end{table}

\textbf{RGBD encoding} has been found to be a modality that adds value to several downstream vision tasks such as classification~\cite{RGBDfusion_old,MVFusionNet}, detection~\cite{RGBD-pretrained}, and especially segmentation~\cite{RedNet,CMX-rgbdfusion,Tokenfusion} where the depth signal is well suited to find and refine contours. Depth information has the potential to extract features invariant to condition/instance-based color artifacts. Hence, yielding a better generalization towards unseen object instances. However, we find that both bi-directional ($c\leftrightarrow d$)~\cite{CMX-rgbdfusion,Symmetric-Cross-modality-Residual-Fusion} and towards color-directed $d \rightarrow c$~\cite{RedNet} fusion is corrupting the RGB encoding (see Tab.~\ref{tab:RGBD}), with the caveat that our findings are on the basis of S.\&E.-based~\cite{Squeeze} RGBD-fusion rather than the direct implementation used by~\cite{CMX-rgbdfusion,Symmetric-Cross-modality-Residual-Fusion,RedNet} in order to yield more general and comparable results. Albeit the usage of further regularization with extra depth noise augmentation and image-wise normalization (removing absolute distances), we find our depth directed fusion ($c \rightarrow d$) implementation to reach at best near-on-par performance with pure MV RGB. It appears that the depth information is at best an opening for overfitting. Related work uses HHA~\cite{HHAorigin} (horizontal, height, angle) encoded depth rather than a simple one-dimensional standardized signal for better results~\cite{HHARGBD}, which we cannot report. However, with our findings regrading pre-trained depth encoding weights~\cite{RedNet}, we argue that the key reason for the RGBD failure is a quality mismatch between color and depth feature extraction. While sophisticated and suitable pre-trained encoders for color images are broadly available, we observe mainly pre-trained depth encoders for segmentation downstream tasks that are fundamentally different. Self-supervised RGBD pre-training methods such as~\cite{RGBDSSL} are promising for efficient training, but not broadly accessible yet, nor does it allow end-to-end training of an RGBD encoder with intermediate RGBD-fusion.  
\begin{table}[h!]
  \begin{center}     
  \begin{tabular}{| l ccccc |} 
    \hline
    Input & Weight Encoding. &Weight Fusion.& Top 1 & Top 3 & Top 5\\    
    \hline \hline
    W & PE + PN & - &64.0 & 88.3 & 97.4\\
    \hline
    2RGB & - & - &94.6&\textbf{99.0}&\textbf{99.5}\\
    2RGBW & PE &Tr-De.&94.9& 98.8 & 99.3\\
    2RGBW & PE+PN &Tr-De.& 94.7 & 98.8 & 99.3\\
    2RGBW & PE+PN' &Tr-De.&\textbf{95.1}& 98.6 & 99.3\\
    \hline
    3RGB & - & - &\textbf{94.6}&98.6&99.1\\
    3RGBW & PE &Tr-De.&94.1&98.2&98.9\\
    3RGBW & PE+PN &Tr-De.& 93.5 & \textbf{98.7}& \textbf{99.2}\\
    3RGBW & PE+PN' &Tr-De.& 93.6 & 98.3 & 98.8\\
    \hline
  \end{tabular}
  \end{center}
  \caption{Results concerning multi modal fusion between MV RGB data and the object weight [kg]. PE denotes the usage of Positional-Encoding~\cite{AttentionIsAll}, while PN denotes a pre-trained network (PropertyNet) for upsampling the PE encodings (PN' is frozen during training).}
  \label{tab:weight_exp}
\end{table}

\textbf{Anchor-based Decoding: } Moreover, we investigate the effect of using weight encoding ($PE(\omega)$) as anchor for the MV-fusion Transformer-Decoder ($\digamma_v$) such that the decoder query $Q = Q_{emb}+ PE(\omega)$ or $Q=WN(PE(\omega))$, where $Q_{emb}$ is a trainable parameter of size $1\times d_d$. However, we can only report marginal gains within different metricise, 
which can be explained by the PropertyNet (PN) accurancy and general weight modality limitation within MVIP. However, anchor-based decoding becomes potentially beneficial when enriching the anchored information (add further property dimensions) or within problems that are not purely solvable by image information (e.g. occluded features/materials which are detectable by physical properties such as weight). 

\section{Conclusion}
We present MVIP, a novel multi-view and multi-modal application oriented dataset for industrial part recognition. The intention of the dataset is to narrow the gap between basic research in ML-based computer vision regarding part recognition and real world industrial applications. MVIP enables researchers to investigate a wide range of open research questions combined in a single benchmark, spanning several domains. Within our experiments, we identify a general lack of strong multi-modal encoders and fusion mechanisms, which can leverage robust and efficient finetuning on industrial downstream tasks. Furthermore, we identify a general risk of overfitting within our end-to-end trained MV problem, which we hypothesize to be related to a statistical increase of MV artifacts and to a low batch size. However, we achieve a baseline Top 1 accuracy of $>95\%$ and Top 5 of $99.5\%$ within MVIP using a MV-RGB classifier due to our novel MH-loss, which we find to yield more stable and better results across the board for MV-fusion.\\ \indent Regarding future work, we aim to investigate the (automatic) generation of synthetic data via MVIP's options for object and scene 3D-reconstruction. Here, one can use MVIP's calibrated RGBD setup to generate textured CAD and point cloud data, which in turn simulations can laverage to create unlimited diverse data in a controlled setup. In addition to classification, these synthetic data generation can also be employed in combination with the NL-tags on object conditions and objects masks to generate data for defect/anomaly and condition detection/classification. Moreover, with MVIP new techniques for MV-augmentation, MV-sampling, and MV-regularization (similar to our MH-Loss) can be investigated that leverage knowledge regarding views, super-classes, and natural language tags. Methods for multi-modal pre-training and fusion between image-based data such as RGB, Depth, and physical properties (width, height, length, weight) as well as natural language tags should can be explored. Other future work can also use MVIP to investigate dataset curation with respect to the selected cameras and the number of images -- aiming for more efficient training and adaptation. Following~\cite{ChavanRealisticIndustrial}, we would also encourage researchers to investigate incremental learning on MVIP, since industrial use cases often have a rapidly changing range of objects. 


\section*{Source code and data}
Please find all the source code related to our experiments, dataset handling, and a download link to MVIP here:\\\\
\textcolor{pink}{Github:} \url{https://github.com/KochPJ/multi-view-part-recognition}. 

\section*{Acknowledgments}
We would like to thank everyone at the Fraunhofer IPK involved in the MVIP project. A special thanks to Vivek Chavan and Maulik Jagtap for their contributions to the dataset design and the digitization.

\bibliographystyle{unsrt}  
\bibliography{references}  


\end{document}